# Merging of Ontologies Through Merging of Their Rules

Olegs Verhodubs

oleg.verhodub@inbox.lv

**Abstract**. Ontology merging is important, but not always effective. The main reason, why ontology merging is not effective, is that ontology merging is performed without considering goals. Goals define the way, in which ontologies to be merged more effectively. The paper illustrates ontology merging by means of rules, which are generated from these ontologies. This is necessary for further use in expert systems.

## I. INTRODUCTION

Continuously developing over the past few decades, the Web today has become a global information resource. With the distributed nature of information as one of its principles, the Web has inherited the advantages and disadvantages associated with this principle. One of the significant advantage of the distributed nature of information in the Web is the variety of this information. A variety of information in the Web is provided by a huge number of people, who supply this information. The main disadvantage of the Web is the information there is situated in different resources. It is necessary to make significant efforts to gather the required information together. An html-based Web was not quite capable of this task. The extension of the html-based Web was developed, and it was named the Semantic Web [1]. It is considered that the Semantic Web is a machine-readable Web that is the Web that can be processed by machines. There are several technologies, aimed at implementing the Semantic Web. These technologies are RDF (Resource Description Framework), RDFS (Resource Description Framework Schema), SPARQL, OWL (Web Ontology Language) and some others [2]. This research focuses on the OWL standard as a standard for describing ontologies. An ontology is a piece of information with its own structure. This structure of ontology is provided by classes, properties, datatypes and individuals [3]. An ontology can describe some strictly defined area, as well as some kind of information resource that accumulates information from different areas. It is foreseen that most ontologies will be generated from websites. Usually no website and ontology generated from the website can fully satisfy the need for information. In this case ontologies should be merged together. Merging of ontologies can be implemented differently. One way is more complete, because merges all elements of two or more ontologies. This is not necessary in any case, therefore sometimes another way is preferable. Another way is more effective, because makes merging process more selective. Only some elements of ontologies are merged. For example, it is possible one ontology to enrich with properties from another ontology. Or it is possible one ontology to enrich with links from another ontology. Or the process of ontology merging can be done somehow differently. One of such ways is shown in the paper. This way foresees to merge functional value of ontologies instead of merging the ontologies themselves. Functional value of ontologies can be different, and it depends on the goal of ontology use. Particularly, ontologies can be used for generation of rules [4], [5]. In this regard functional value of ontologies are rules, which are generated from ontologies. Merging of rules, generated from different ontologies, is merging of ontology functional value and can be called as functional merging of ontologies. Functional merging of ontologies is described in the paper.

This paper is divided into several sections. The next section reviews previous researches in the area. The third section describes merging of ontologies throug the rules, generated from these ontologies. In turn, the fourth section presents the architecture of merged ontologies. The last section introduces the conclusions of the research, and also research directions for the future.

**II. PREVIOUS RESEARCH**

Integration of ontologies is one of the significant tasks in the area of the Semantic Web. That is why there are a lot of papers dedicated to the subject. Some overview of ontology integration methods was done in [6]. Several years have passed since publication, but the overview of ontology merging methods in the paper is still relevant. Although mentioned overview is performed for the Semantic Web Expert System [7], it is also useful for other researches.

So, there are ontology mapping, ontology alignment and ontology merging [6]. Ontology mapping is a specification of the semantic overlap between two or more ontologies. Ontology alignment is creating links between two original ontologies. Ontology merging is the process of a new ontology creation, which is the union of one or more source ontologies. Ontology merging means that a single ontology is generated from two or more ontologies. Certainly, each type of ontology integration has its own advantages and disadvantages. Only disadvantages are reviewed here, because advantages are due to existence of these types of ontology integration, but disadvantages determine why the types of integration of ontologies are not suitable for us. The semantic overlap between two or more ontologies in ontology mapping implies some measure of fuzziness and incompleteness. The subprocess of creating links between two original ontologies during the process of ontology alignment implies the creation of something like a virtual integrated ontology. The weak spots here are links, which can break off completely or temporarily. Ontology merging is critical to memory, because generation of a single ontology from two or more source ontologies can require a large amount of memory. This may occur if source ontologies are large or the process of ontology merging includes merging of more than two ontologies.

Thus, no one type of ontology integration satisfies, considering mentioned disadvantages. It is necessary to take into account the purposes of ontology use in order to develop the alternative type of ontology integration without obvious disadvantages.

**III. MERGING OF ONTOLOGIES VIA THEIR RULES**

There are a lot of ontology uses. For example, ontologies can describe a particular field of human activity. It could be sports, medicine or something else. Or ontologies can be used for transforming the Web towards to be more processable by machines. For instance, it is possible to process ontologies by means of Apache Jena Inference Api [8]. In any case, data, information and knowledge from different ontologies can intertwine with each other giving rise to the need for ontology merging. The difference and similarity of data, information and knowledge is shown in [9]. Here is important that ontologies in themselves can provide us with knowledge namely rules if production systems are meant [4], [5]. The purpose is to use knowledge that is rules generated from ontologies in expert systems. The algorithm of ontology use in expert system is the following. First of all, rules are generated from ontology. Then the generated rules are supplied to the knowledge base of the expert system. After that, rules from the knowledge base can be used in the expert system. If the rules, which are generated from one ontology, are not enough, then several ontologies are merged into a single ontology and only after that the rules are generated from a single, integrated ontology. This way has at least one serious disadvantage: merging of two or more ontologies is a long process. The process of rule generation from a single, merged ontology is a long process, too. Both of these processes can significantly slow down the work with the expert system and thereby reduce its attractiveness to a critical level. It would be possible to merge pairs, triples or etc. of ontologies in advance, but what exactly ontologies to merge becomes known only from user's request, that is, it is impossible to know it in advance. However, some of the work of merging ontologies should be carried out in advance as it is realized in the search engines, when web pages are indexed in advanced, and the user actually utilizes the database, when types his keywords in the intended for this string. That is, the search in the Web when working with some of the search engines does not carry out in the moment of typing keywords; in the moment of typing the search is carrying out in the indexed database.

It is possible to merge the ontologies in a special way. Not the ontologies themselves, but the product of their processing should be merged for this purpose. Rules, which are generated from ontologies, are the product of ontology processing. In this way, ontologies are being merged by means of rule integration together. That is, rules, which are generated from one ontology, and rules, which are generated from another ontology, are used together if they would be generated from one, single ontology. This is different from merging of ontologies in the usual sense of the word, when parts of one ontology are being added to another ontology [6].

Advantages and disadvantages are reviewed further in order to evaluate the proposed way of ontology merging. The main advantages are getting rules from being processed ontologies and possibility to realize the process of rule generation from ontologies, before these rules are needed in the expert system. Early ontology processing positively affects the speed of rule use, because time is needed to access the database, where generated rules are stored. It is possible to suppose that in such a way the speed of the expert system is approximately equal to the speed of the web search engine. In turn, two principal disadvantages of this ontology merging are observed. The first one is the single ontology from several other ontologies is not generated as it happens during usual ontology merging. This is not the purpose; however the enrichment of ontologies by parts of other ontologies can be useful from the optimization point of view. The second disadvantage is not so evident, but more important for quality of the generated rules. The fact is that the rules, which are generated from each ontology separately, can differ from the rules, which are generated from the single merged ontology, which is generated from several other ontologies. Let us demonstrate the situation (Fig.1.):

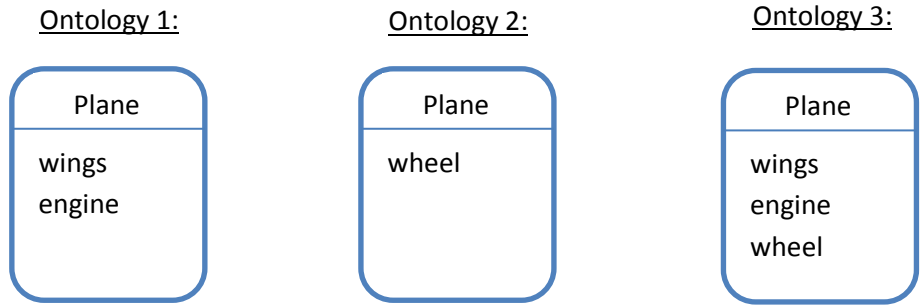

Fig.1. Three ontologies with class "Plane".

There are three ontologies (Fig.1): "Ontology 1", "Ontology 2" and "Ontology 3". "Ontology 1" and "Ontology 2" are two different ontologies, which contain the class "Plane". The class "Plane" in the "Ontology 1" has two properties "wings" and "engine". The class "Plane" in the "Ontology 2" has one property "wheel". "Ontology 3" is the result of "Ontology 1" and "Ontology 2" merging. The class "Plane" in the "Ontology 3" has three properties: "wings", "engine" and "wheel". So, it is possible to generate the following rules from these ontologies according to [4]:

TABLE I. Ontologies and generated rules.

| Ontology | Rules |
| --- | --- |
| Ontology 1 | **IF** *wings* **and** *engine* **THEN** *Plane* |
| Ontology 2 | **IF** *wheel* **THEN** *Plane* |
| Ontology 3 | **IF** *wings* **and** *engine* **and** *wheel* **THEN** *Plane* |

It can be found that the rules, generated from "Ontology 1" and "Ontology 2" differ from the rule, which is generated from the "Ontology 3". The rule from the "Ontology 3" is more complete and is preferred for expert systems. Thus, if ontologies are being merged by means of the rules, which are generated from ontologies and collected together, some extra work is needed for increasing the quality of the generated rules.

**IV. ARCHITECTURE FOR ONTOLOGY MERGING**

Merging of ontologies in the form of a rule collection is necessary for functioning of the expert systems. Considering a lot of rule types, which can be generated from the ontology [4] [5], and also the amount of ontologies, which can be involved in the process of rule generation, the architecture of the storage for rule storing and retrieving has to be developed. It is possible to work out your own software for storing and retrieving rules in the expert systems or to utilize ready-made software for mentioned purposes. DBMS (Data Base Management System) is software that usually is being used for such tasks. DBMS had already been optimized in many parameters, therefore DBMS use is preferred. There are a lot of DBMS that differ by its type, prevalence, licensing and so on. MySQL [10] DBMS is quite sufficient for research tasks as open-source, quite effective, well-documented and tested in real tasks, that is why MySQL is chosen.

Architecture of DB (Data Base) affects the efficiency of working with data, therefore it is necessary to pay close attention to this task. The DB architecture is stored data, relationship of this data and kind of grouping this data, and the DB architecture is set using the DBMS means.

MySQL DBMS is a relational DBMS [11]; this means that data in the DB is grouped by tables, but tables are interconnected by means of primary and foreign keys. So, it is necessary to define the types of data, number of tables and their structure, and also the links between the tables of DB in order to develop the architecture of rule storage. A rule consists of the condition and the result. In addition, the condition and the result of every rule both have its own membership functions. Thus, the table for storing rules in the DB has the following structure (TABLE II):

TABLE II. Table for storing rules.

| id | Condition | MFC | Result | MFR |
|---|---|---|---|---|

Let us explain the abbreviations (TABLE II): id is an index number of the record (primary key), condition is a condition of the rule, MFC is a membership function of the condition, result is a result of the rule, MFR is a membership function of the rule. id is an integer, Condition and Result are strings, MFC and MFR are real numbers.

Ontologies in the Web can be modified, therefore the rules, which are generated from the same ontology in different times can differ. That is why it is necessary to manage the information about the ontologies, which are participated in the process of rule generation. This may be useful to build automatic system for updating rules in the DB. The table for storing information about used ontologies is the following (TABLE III):

TABLE III. Table for storing information about ontologies.

| id | Name | Address | Access_date |
|---|---|---|---|

Table for storing information about ontologies has the following structure (TABLE III): id is an index number of the record (primary key), Name is a string for storing the name of ontology, Address is a string for storing the address of the ontology in the Web, Access_date is a date, when ontology is accessed and processed.

It is necessary to modify the table for storing rules (TABLE II) in order to add to the each rule information about the ontology, from where it is generated. In this regard, the modified table for storing rules is the following (TABLE IV):

TABLE IV. Modified table for storing rules.

| id | Ontology$^*$ | Condition | MFC | Result | MFR |
|---|---|---|---|---|---|

Modified table for storing rules (TABLE IV) differs from the original table (TABLE II) by the presence of Ontology$^*$ column that is foreign key, which is primary key of the table for storing information about ontologies (TABLE III).

So, there is a table for storing information about all processed ontologies (TABLE III) and a table for storing all generated rules from the processed ontologies (TABLE IV). One more table is necessary for storing the information about all mergers of ontologies. The structure of the table for storing the information about all mergers of ontologies is the following (TABLE V):

TABLE V. Table for storing information about mergers of ontologies.

| id | Table name | Key words | Merged ontologies |
|---|---|---|---|

The table for storing information about mergers of ontologies has the following structure (TABLE V): id is an index number, Key words define the domain of the merged ontologies, Merged ontologies is a list of ontologies that are merged, Table name is a name of the separate table, which contains all rules from merged ontologies. The structure of such a table can be identical to the mentioned table (TABLE II).

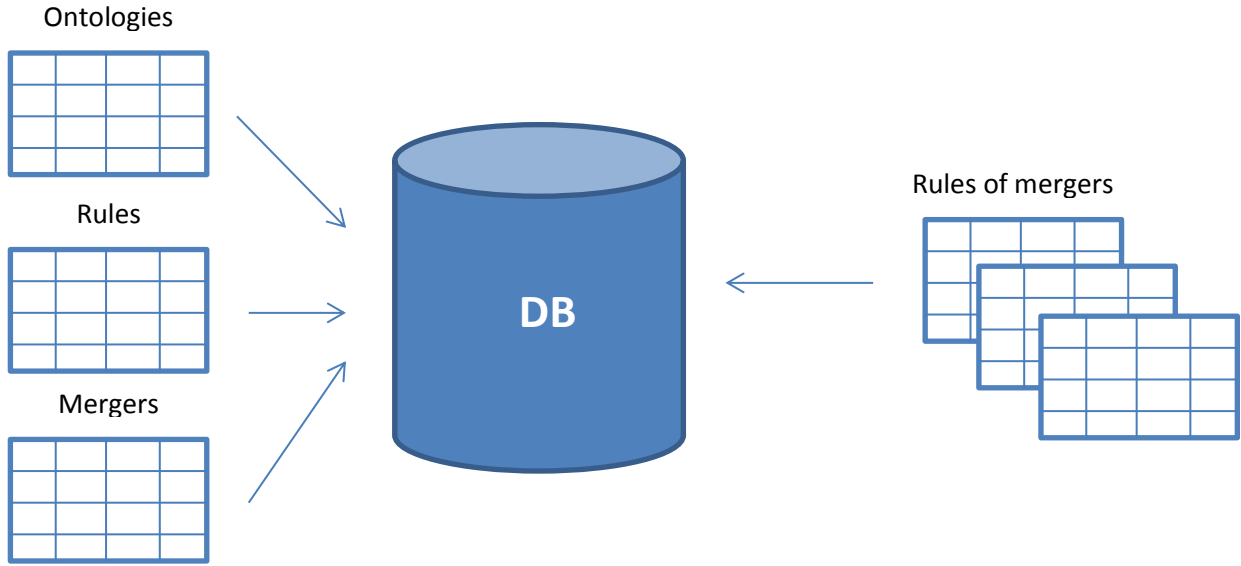

Fig.2. Architecture of DB for ontology merging.

Thus, the architecture of DB to provide ontology merging via rules, generated from these ontologies, consists of two conditional parts (Fig.2). The first part is a structure of the DB; it is provided with three tables: Ontologies, Rules and Mergers. Here the table "Rules" stores all rules from all ontologies, the table "Ontologies" stores information about all ontologies, the table "Mergers" stores information about ontologies to be merged. The second conditional part of the DB consists of the tables that are needed for storing information about each concrete merger of two or more ontologies. Unlike the first conditional part of the DB, which consists strictly from three tables, the amount of tables here is not constant and depends on the amount of ontology mergers. For example, if there are twenty mergers of ontologies, then twenty tables are in the second part of the DB.

**V. CONCLUSION**

The paper describes one more way to merge the ontologies. Merging of ontologies is possible to perform if to be guided by the purpose, for which merging of ontologies is needed. Functioning of the expert systems, based on rules, is such a purpose. Merging of rules, which are generated from several ontologies, is the same as merging of ontologies in terms of use of ontology merging in expert systems. Merging of rules is grouping the rules in order to use these rules in the expert system according to the user's request.

The paper presents the DB architecture for soring rules, generated from ontologies, and mergers of ontologies. This information in the DB is needed for functioning of expert systems. It is important to mention that such an architecture of the DB is sufficient for working of the expert systems, but this architecture is not ideal. Possible improvement of the DB architecture is associated with the realization that is why it is not discussed here.

This research identified such an interesting task: rules, generated from each separated ontology, differed from the rules, generated from one, single merged ontology. Single merged ontology here is the ontology that is merged from several other ontologies in usual way that is parts of several ontologies are united in one [6]. It would be supposed that processing of the rules, generated from several ontologies, is useful and has good research potential.

## ACKNOWLEDGEMENTS

This work like most previous works has been supported by my family and my friends.